%% file: leapwing-report.tex
\documentclass[10pt]{article}
\PassOptionsToPackage{table}{xcolor}
\usepackage{leapwing}
\usepackage{nicefrac,algorithm,algorithmic,multirow,subcaption,wrapfig}
\colorlet{LightGray}{LWSurface}

\newcommand{\eqcontrib}{\ensuremath{\dagger}}
\newcommand{\projectlead}{\ensuremath{\ddagger}}
\newcommand{\eqsupervision}{\S}

\newcommand{\answerTODO}[1][]{\textcolor{red}{\textbf{[TODO]}}}
\renewcommand{\LWPartnerLogoHeight}{11mm}
\renewcommand{\LWPartnerLogos}{%
  \LWPartnerLogo[12mm]{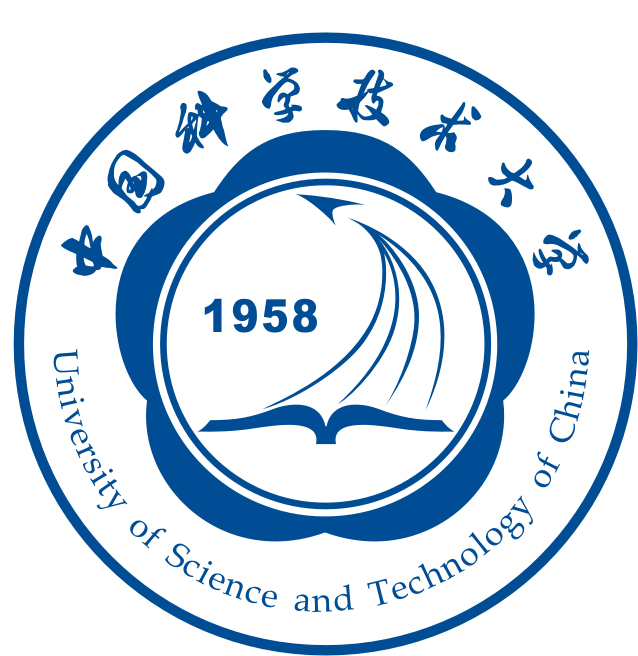}
}
\renewcommand{\LWDate}{August 2026}
\newif\ifLWChecklist
\LWChecklisttrue
\renewcommand{\LWTitle}{Seeing Together, Acting Apart:\\Shared Environmental Understanding\\for Multi-Robot Navigation}
\renewcommand{\LWShortTitle}{SEER: Shared Environmental Understanding}
\renewcommand{\LWReportID}{Technical Report}
\renewcommand{\LWAuthors}{%
Haihong Hao$^{1\eqcontrib}$\quad Lei Chen$^{1\eqcontrib}$\quad Mingfei Han$^{1,2,3\eqcontrib\projectlead}$\quad Chenglong Yan$^{1}$\quad Dong An$^{3}$\\[2pt]
Yuqiang Yang$^{4}$\quad Changlin Li$^{5}$\quad Minghao Guo$^{3}$\quad Salman Khan$^{3\eqsupervision}$\quad Xiaojun Chang$^{1,2\eqsupervision,*}$}
\renewcommand{\LWAffiliations}{%
$^{1}$Future Media Computing Lab, USTC\quad $^{2}$LeapWing AI\quad $^{3}$MBZUAI\quad
$^{4}$Shanghai AI Laboratory\quad $^{5}$Stanford University}
\renewcommand{\LWNotes}{$^{\eqcontrib}$ Equal contribution.\quad $^{\projectlead}$ Project lead.\quad $^{\eqsupervision}$ Equal supervision.\quad $^{*}$Corresponding author.}
\renewcommand{\LWLinks}{}
\renewcommand{\LWAbstract}{\begingroup
\renewenvironment{abstract}{}{}\input{sec/0_abs}\endgroup}
\begin{document}
\makeLWTitle
\input{sec/1_intro}

\input{sec/2_related_works}
\input{sec/3_methods}

\input{sec/4_experiments}
\section{Conclusion}
This paper argues for a shift in how multi-robot navigation should be understood. The essence of collaboration is not merely that multiple robots move at the same time, avoid each other, or exchange action histories; it is that each robot can make the world more legible for the others. By formulating multi-robot VLN around shared environmental understanding, we move the problem from surface-level coordination to a deeper question of collective scene comprehension. Complementary embodied experience can be transformed into shared knowledge that improves navigation beyond isolated single-agent perception. The real-robot study further shows that this idea is not confined to benchmark artifacts. We believe this representation-centric view offers a cleaner, more scalable, and more general foundation for embodied multi-agent systems.
\bibliographystyle{unsrtnat}
\bibliography{references}
\appendix
\input{sec/appendix}
\end{document}

%% file: sec/0_abs.tex
\begin{abstract}

Multi-robot embodied systems are typically designed to collaborate by coordinating actions, sharing trajectories, or learning joint policies. We ask a complementary question: can collaboration emerge before decision-making, by first building a shared understanding of the environment?
We introduce \emph{Shared Environmental Understanding}, a policy-agnostic representation that aggregates partial egocentric observations from multiple robots into a shared latent world model. This shifts collaboration from the policy space to the representation space, enabling decentralized decision-making with shared environmental context.
To this end, we propose \textbf{S}hared \textbf{E}nvironment \textbf{E}ncoding and \textbf{R}etrieval~(SEER), a framework that organizes multi-robot observations into a shared memory, retrieves relevant context via spatio-temporal similarity, and encodes it using a predictive latent world model. The resulting representations are injected into downstream navigation agents as tokens. 
Experimental results demonstrate that SEER captures action-relevant environmental structure and consistently improves navigation performance across both specialized VLN agents and general vision-language model agents, while preserving single-agent capabilities.
\end{abstract}

%% file: sec/1_intro.tex
\section{Introduction}

Multi-robot embodied systems are commonly expected to collaborate by coordinating their behaviors: agents may communicate, share experience, divide tasks, avoid conflicts, or jointly decide what to do next.
This question naturally extends classic embodied navigation and vision-and-language navigation settings, where an agent must ground observations and instructions into actions in realistic environments~\citep{anderson2018vision,hong2021vlnbert}. 
It has also motivated a large body of work on multi-robot coordination, communication, task allocation, and multi-agent decision making~\citep{dias2006market,yan2013survey,sukhbaatar2016learning,lowe2017multi}, as well as recent embodied multi-agent systems that share semantic maps, messages, frontier assignments, action plans, or generated rollouts for cooperative navigation and planning~\citep{liu2022multiagent,yu2023conavgpt,zhang2025combo}. 
However, most existing approaches frame collaboration at the level of agent-centric traces, such as sharing actions, trajectories, maps, or intermediate decisions. While effective, these methods tightly couple collaboration with policy design and communication protocols. This raises a fundamental question: can collaboration emerge from shared perception before shared decision-making? In other words, instead of exchanging what agents \emph{do}, can they first share what they \emph{understand} about the environment?

In realistic settings, different robots naturally observe complementary parts of the same environment from diverse viewpoints and at different times. Aggregating these partial observations can yield a richer representation of the environment than any single agent can obtain alone. Crucially, such a representation can serve as a policy-agnostic interface: it can be shared across agents while allowing each agent to make decisions independently using its own policy.  We therefore propose to shift collaboration from policy space to representation space. We introduce \emph{Shared Environmental Understanding}, a policy-agnostic representation that aggregates multi-robot observations into a shared latent description of the environment. This shared representation provides a common environmental context, while allowing each agent to act independently with its own policy.

In this paper, we propose \emph{Shared Environment Encoding and Retrieval} (SEER), a framework that learns shared latent world representations from multi-robot, multi-view observations. SEER organizes distributed observations into a shared memory, retrieves relevant context via spatio-temporal similarity, and encodes it using a predictive latent world model. The resulting representations are exposed as continuous tokens and injected into downstream navigation agents, enabling decentralized decision-making with shared environmental context.
We evaluate SEER from both representation and task perspectives. Our results show that the learned representations capture action-relevant environmental structure and consistently improve downstream navigation across both specialized VLN agents and general vision-language model agents.
\begin{figure}[t]
    \centering
    \includegraphics[width=\textwidth]{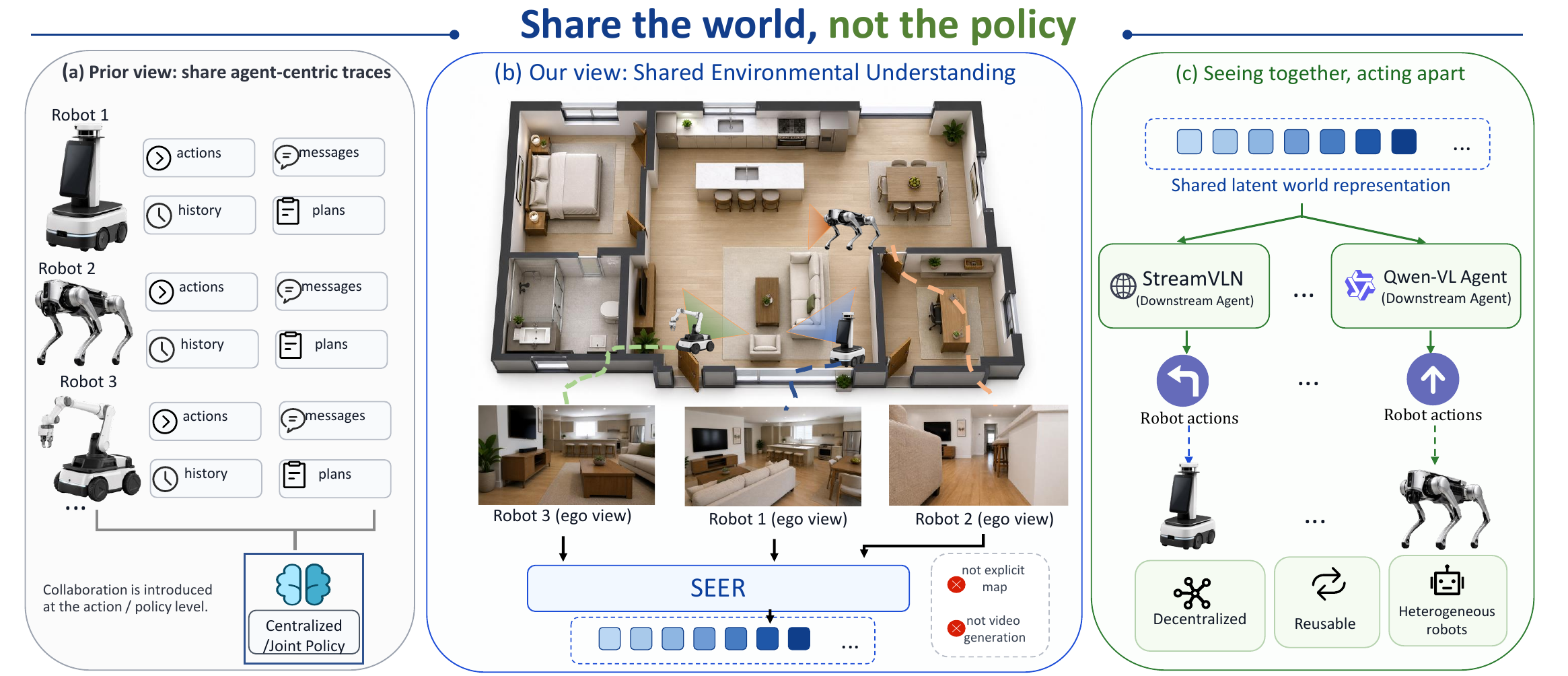}
\caption{
Unlike prior methods, SEER builds a policy-agnostic shared world representation by pooling partial egocentric observations, shift collaboration from policy space to representation space. 
}
    \label{fig:seer_motivation}
\end{figure}

%% file: sec/2_related_works.tex
\section{Related Works}

\textbf{Multi-Robot Coordination.} Multi-robot coordination covers task allocation, communication, distributed planning, heterogeneous cooperation, distributed optimization, coordinated exploration, mapping, and aerial/multi-UAV swarms~\citep{dias2006market,yan2013survey,rizk2019cooperative,halsted2021survey,burgard2005coordinated,chung2018swarm}. Learning-based approaches study inter-agent communication and decision making through differentiable communication and MARL~\citep{sukhbaatar2016learning,lowe2017multi,rashid2018qmix,das2019tarmac,iqbal2019actor}, while recent embodied systems extend coordination to cooperative visual-semantic navigation, ambiguous instruction resolution, heterogeneous collaboration, LLM/VLM-driven planning, and embodied multi-agent benchmarks~\citep{liu2022multiagent,liu2022ambiguous,liu2023heterogeneous,yu2023conavgpt,kannan2023smartllm,chang2025partnr,zhang2025combo}. Related ideas also appear in cooperative driving/perception, city-scale simulation, and multi-view-consistent generation for urban or driving worlds~\citep{wang2020v2vnet,xu2022v2xvit,xu2023cobevt,park2023generativeagents,gao2024embodiedcity,bougie2025citysim,shi2023mvdream,liu2024syncdreamer,wen2024panacea,jiang2025dive}. \textbf{World Models for Embodied Agents.} Embodied world modeling builds on large-scale datasets, simulators, and 3D language-grounding benchmarks, including Matterport3D, ScanNet, Habitat, HM3D-Sem, EmbodiedScan, EmbodiedQA, ScanRefer, ReferIt3D, ScanQA, SQA3D, OpenEQA, 3D-LLM, and SIG3D~\citep{chang2017matterport3d,dai2017scannet,savva2019habitat,yadav2023hm3dsem,wang2024embodiedscan,chen2020scanrefer,achlioptas2020referit3d,azuma2022scanqa,majumdar2024openeqa,huang2024leo,hong2023threedllm,man2024sig3d}. World models have evolved from latent dynamics and imagination-based control~\citep{ha2018worldmodels,hafner2019planet,hafner2023dreamerv3,wu2023daydreamer} to interactive/generative prediction of future observations, occupancy, and navigation rollouts~\citep{bruce2024genie,wang2024worlddreamer,hao2024hierarchical,zheng2024occworld,bar2025navigation}; representation-predictive methods such as V-JEPA and V-JEPA 2 further highlight latent-feature prediction over pixel reconstruction~\citep{bardes2024revisiting,assran2025vjepa2}. Recent embodied models also encode explicit 3D/4D structure or navigation geometry and connect perception with language and action~\citep{lei2024gaussnav,zhen2025tesseract,brohan2023rt2,kim2024openvla,black2024pi0,zhao2025cotvla,ma2024vlasurvey,guo2026phaselora}. \textbf{Vision-and-Language Navigation.} VLN grounds natural-language instructions into sequential navigation actions. Early work established standard benchmarks and agents, including speaker-follower training, cross-modal matching, self-monitoring, regretful navigation, and back-translation~\citep{anderson2018vision,fried2018speaker,wang2019rcm,ma2019selfmonitoring,ma2019regretful,tan2019backtranslation}. The field later expanded to multilingual, dialog, object-centric, continuous, and outdoor settings~\citep{ku2020rxr,qi2020reverie,zhu2021soon,krantz2020vlnce,thomason2020cvdn,chen2019touchdown}, and recent agents use large-scale pretraining, transformers, map-centric or history-aware representations, future-view prediction, VLM/LLM reasoning, map-guided planning, causal learning, aerial VLN, long-horizon and streaming settings, physical deployment, semantic-map memory, and world-model-based navigation~\citep{hao2020prevalent,hong2021vlnbert,chen2021hamt,chen2022duet,an2023bevbert,li2023futureview,liu2023aerialvln,zhou2024navgpt,chen2024mapgpt,wang2024causal,song2025lhvln,wei2025streamvln,wang2025vlnpe,zhang2025citynavagent,zhang2025mapnav,zhang2025vlfly,yao2025navmorph,hao2026latentpilot,han2026implicit,hao2026conav,shao2026baldro,zhai2026maximizing}. In contrast to prior coordination works that introduce controllers or communication protocols, world models that primarily focus on single-agent prediction, and VLN methods that mainly improve policies, memory, prompting, or deployment, our work learns a policy-agnostic shared environmental representation from complementary multi-robot observations and uses it as reusable context for downstream navigation agents.

%% file: sec/3_methods.tex
\begin{figure}[t]
    \centering
    \includegraphics[width=\textwidth]{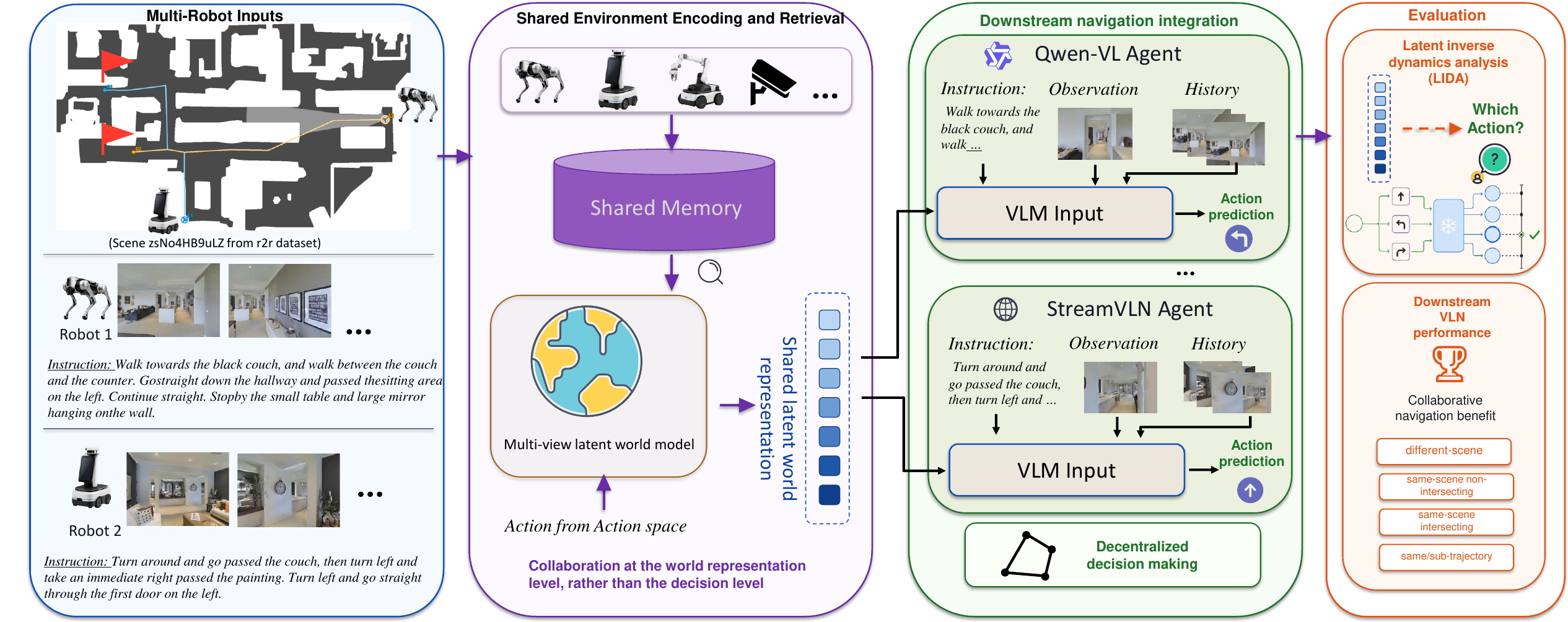}
\caption{
Our SEER organizes distributed observations into a shared memory, retrieves relevant context via spatio-temporal similarity, and encodes it using a predictive latent world model.
}
\label{fig:overall_pipeline}
\end{figure}

\section{Method}  
We propose a framework that leverages shared environmental understanding to learn policy-agnostic representations. The resulting shared environmental tokens are injected into downstream navigation agents, enabling both representation probing and evaluation on VLN tasks (Figure~\ref{fig:overall_pipeline}).


\subsection{Problem Definition}

In conventional multi-robot navigation, consider $N$ robots navigating in the same environment $\mathcal{E}$. 
At time step $t$, robot $i$ receives a partial egocentric observation
$
    o_t^i \in \mathcal{O},
$
executes an action
$
    a_t^i \in \mathcal{A},
$
and robot also be conditioned on a language instruction
$
    x^i \in \mathcal{X}.
$
The trajectory of robot $i$ is denoted as
$
    \tau^i = \{(o_t^i, a_t^i)\}_{t=1}^{T_i},
$
then the multi-robot observation data is collected as 
$
    \mathcal{D}_{\mathcal{E}} = \{\tau^1, \tau^2, \ldots, \tau^N\}. 
$
Since each $o_t^i$ only captures a local egocentric view, no single robot can fully observe the environment. 
However, the set of observations from different robots,
\begin{equation}
    \mathcal{O}_{\leq t}^{1:N} = \{o_s^i \mid i=1,\ldots,N,\; s \leq t\},
\end{equation}
provides complementary multi-view observations about the same underlying world.
We define \emph{Shared Environmental Understanding} (SEU) as a policy-agnostic environmental representation that aggregates partial egocentric observations from multiple robots into a common representation, which can be reused by different downstream navigation agents without enforcing a centralized controller. 
Concretely, we aim to learn a function
\begin{equation}
    \mathbf{z}_t^{\mathrm{SEU}}
    =
    f_{\theta}\!\left(\mathcal{O}_{\leq t}^{1:N}\right),
\end{equation}
where $\mathbf{z}_t^{\mathrm{SEU}}$ summarizes multi-robot partial observations into a policy-agnostic environmental representation.
A downstream navigation agent can then make decisions using its own local observation, instruction, and the shared environmental context:
$
    a_t^i \sim \pi_i\!\left(a \mid o_t^i, x^i, \mathbf{z}_t^{\mathrm{SEU}}\right).
$
Importantly, $\pi_i$ can be different across robots or downstream models, meaning that \emph{Shared Environmental Understanding} provides shared environmental information without requiring all robots to share the same policy.

\subsection{SEER: Shared Environment Encoding and Retrieval}
We implement \emph{Shared Environmental Understanding} through \emph{Shared Environment Encoding and Retrieval} (SEER). SEER consists of three key components. First, \emph{SEER Pool} is an identity-agnostic memory that aggregates observations or latent features across robots. Second, \emph{SEER Retrieval} selects relevant context from the pool via spatio-temporal similarity-based sampling, enabling effective information sharing. Third, \emph{SEER-World} is a multi-view latent world model that encodes fragmented egocentric observations into a shared environmental representation.

\begin{figure}[t]
    \centering
    \includegraphics[width=1\textwidth]{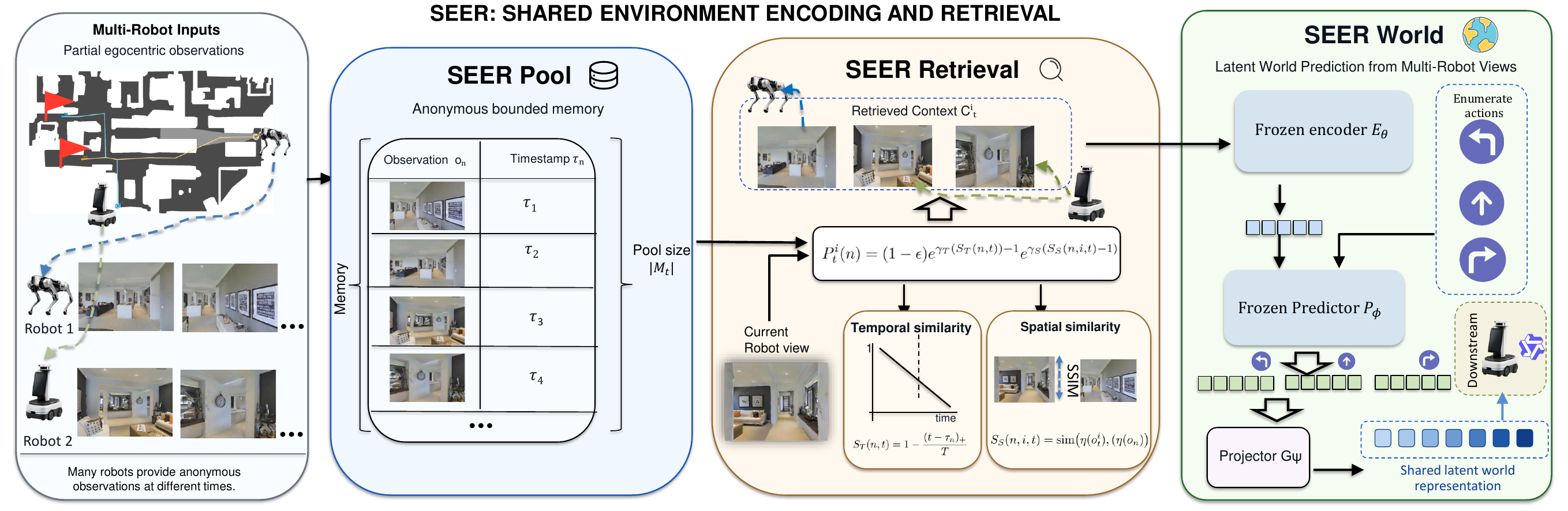}
\caption{\small{Overview of our shared Environment Encoding and Retrieval with SEER-World.
}}
\label{fig:seer}
\end{figure}

\subsubsection{SEER Pool and SEER Retrieval}
\begin{wrapfigure}{r}{.40\textwidth}
    \centering
    \captionsetup{font=footnotesize,skip=4pt}
    \includegraphics[width=\linewidth]{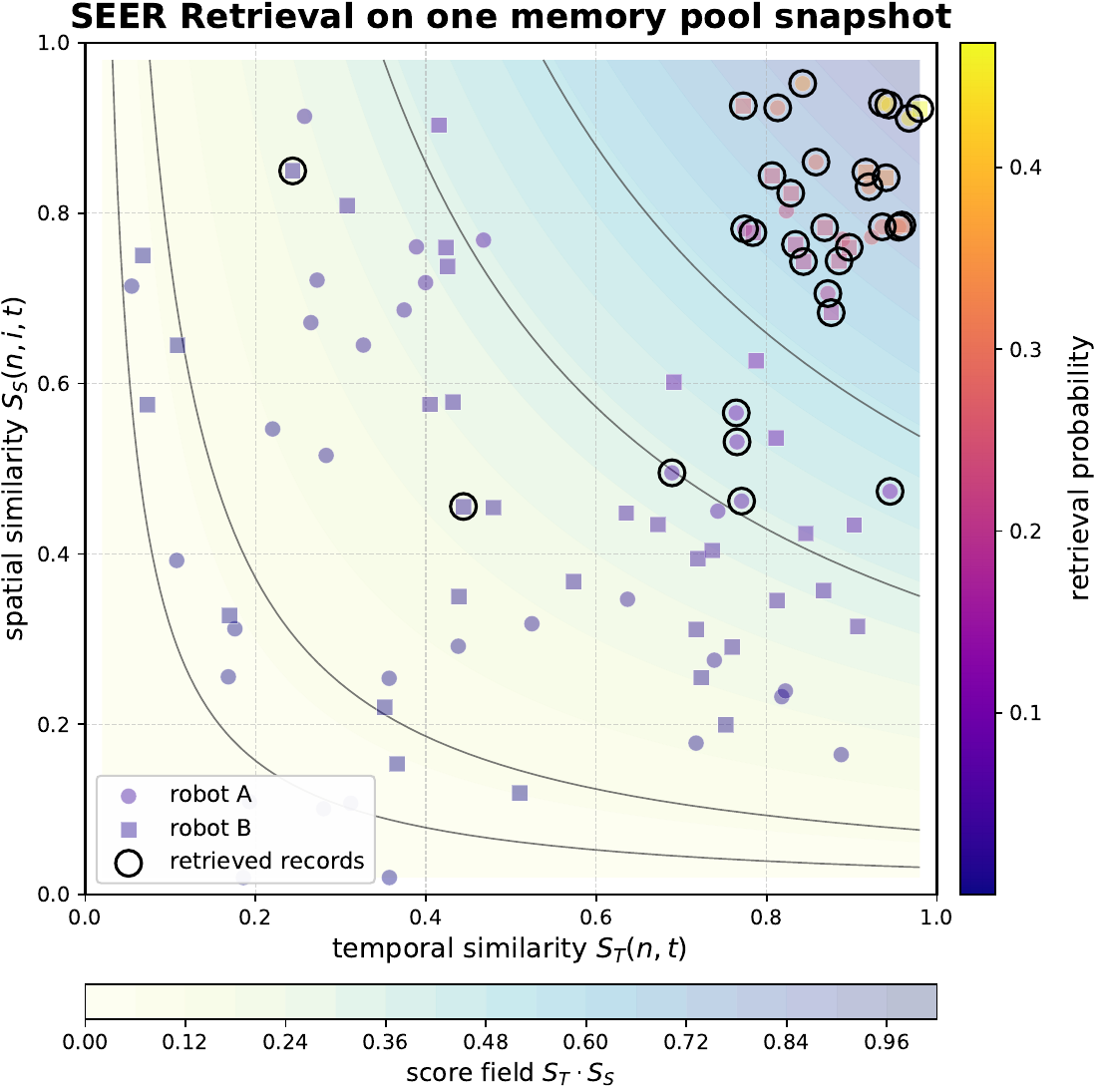}
    \caption{SEER Pool and Retrieval.}\label{fig:seer-retrieval}
\end{wrapfigure}


Different robots observe the same environment from different egocentric viewpoints and at different time steps. 
Naively concatenating all observations would be inefficient and would introduce many irrelevant views. 
To address this issue, SEER introduces an anonymous shared memory mechanism, consisting of the \emph{SEER Pool} and \emph{SEER Retrieval}. 
The goal is to store multi-robot observations as shared environmental context, and retrieve only the observations that are likely to be useful for the current robot under a fixed context budget.

At time $t$, the shared memory pool \emph{SEER Pool} is denoted as
$
    \mathcal{M}_t
    =
    \{m_n\}_{n=1}^{|\mathcal{M}_t|},
$
where $n$ indexes memory entries. Each record $m_n$ contains an observation $o_n$ and its timestamp $\tau_n$. The bounded pool size prevents unstructured growth as more robots contribute observations.

To retrieve useful context from $\mathcal{M}_t$ under a fixed budget, we introduce a time-space-aware retrieval probability, inspired by budget-aware temporal sampling in NavFoM~\cite{zhang2025embodied}. This mechanism prioritizes observations that are both temporally recent and spatially relevant to the querying robot.

Given the current observation $o_t^i$ of robot $i$, we assign each memory record $m_n$ a spatio-temporal weight and normalize it into a sampling probability:
\begin{equation}
\begin{aligned}
w_t^i(n)
&=
(1-\epsilon)\exp\!\big(\gamma_T(S_T(n,t)-1)\big)
\exp\!\big(\gamma_S(S_S(n,i,t)-1)\big)+\epsilon,\\
P_t^i(n)
&=
\frac{w_t^i(n)}
{\sum_{m=1}^{|\mathcal{M}_t|}w_t^i(m)},
\qquad 0<\epsilon<1.
\end{aligned}
\end{equation}
We use weighted sampling without replacement, renormalizing over the remaining records after each draw. All main experiments use a FIFO pool of 1,024 records, a total context budget of $K=8$ observations, $\gamma_T=\gamma_S=1$, and $\epsilon=0.1$. Missing cross-robot context is supplemented with recent self-history. Only observations with $\tau_n\leq t$ are eligible; same-timestep observations from other robots become available once uploaded, while future observations are excluded.

The temporal similarity $S_T(n,t)$ is defined as
\begin{equation}
S_T(n,t) = 1 - \frac{(t - \tau_n)_+}{T},
\end{equation}
where $\tau_n$ is the timestamp of memory record $m_n$, $(\cdot)_+$ denotes the positive part, and $T$ is the temporal horizon controlling the decay.
The spatial similarity $S_S(n,i,t)$ measures the relevance between the current observation $o_t^i$ and the stored observation $o_n$:
\begin{equation}
S_S(n,i,t)=\mathrm{sim}\!\left(\eta(o_t^i), \eta(o_n)\right),
\end{equation}
where $\eta(\cdot)$ denotes the visual encoder and $\mathrm{sim}(\cdot,\cdot)$ is implemented as channel-wise SSIM. SSIM used as a lightweight structural correlation for coarse spatial layout agreement under a fixed retrieval budget, inspired by previous work~\cite{wang2004image,kobayashi2016structured,wentzel2020cohort,zhang2024deepssim}.
Both $S_T(n,t)$ and $S_S(n,i,t)$ are normalized to $[0,1]$, with larger values indicating higher temporal recency and spatial relevance, respectively. The coefficients $\gamma_T$ and $\gamma_S$ control the relative contributions of temporal and spatial similarity in the retrieval process.


Under a fixed context budget, SEER Retrieval samples a compact set of observations from $\mathcal{M}_t$ according to $P_t^i(n)$, without replacement. 
The sampled observations are then ordered by timestamp and combined with the current observation to form the context sequence $ \mathcal{C}_t^i$.
This context is then passed to our SEER-World~$F_{\theta,\phi}$.

This retrieval formulation exhibits several desirable properties. 
First, the probability remains strictly positive due to the lower-bound term $\epsilon$, ensuring that older or less similar observations are not entirely discarded.
Second, the temporal factor $\exp\!\big(\gamma_T (S_T(n,t)-1)\big)$ biases sampling toward recent observations, preserving short-term context critical for navigation.
Third, the spatial factor $\exp\!\big(\gamma_S (S_S(n,i,t)-1)\big)$ suppresses temporally recent but irrelevant views, while promoting observations that are spatially or semantically aligned with the current input.
Fourth, the fixed context budget ensures that the number of retrieved tokens does not scale with the number of robots, viewpoints, or stored observations.
Finally, by operating over an anonymous memory pool, SEER naturally generalizes to varying numbers of robots and viewpoints: additional robots enrich $\mathcal{M}_t$, while each agent still retrieves a bounded, relevance-weighted context.

\subsubsection{SEER-World: Latent World Prediction from Multi-Robot Views}
\begin{wrapfigure}{r}{.48\textwidth}
    \centering
    \captionsetup{font=footnotesize,skip=4pt}
    \includegraphics[width=\linewidth]{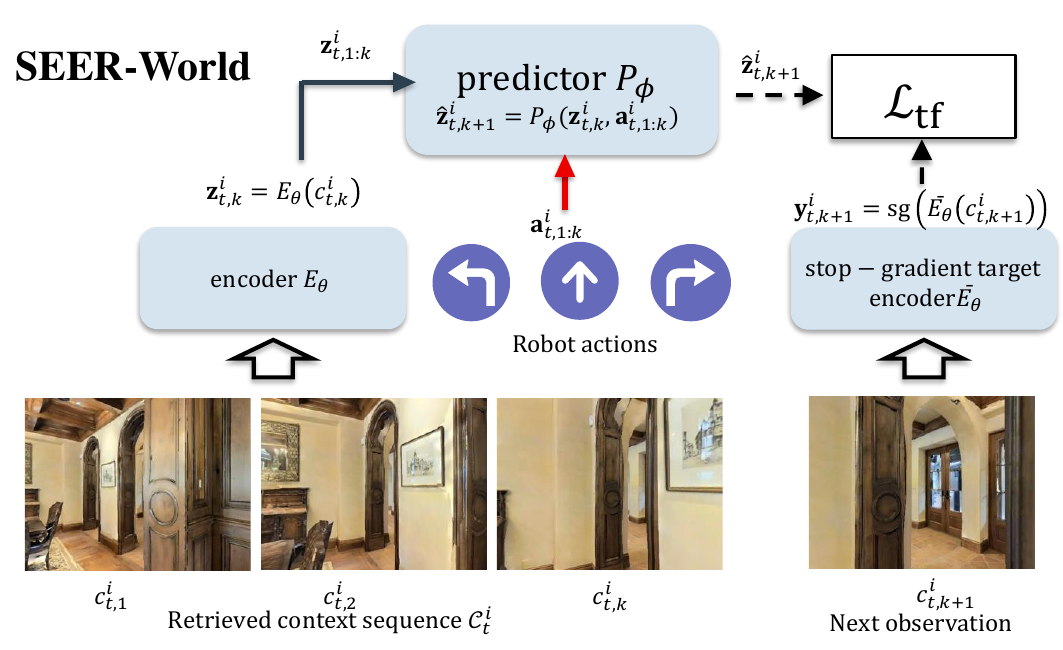}
    \caption{SEER-World architecture.}\label{fig:seer-world-structure}
\end{wrapfigure}
SEER-World converts retrieved multi-robot observations into a shared environmental representation by predicting future states in latent space.
This design follows the central motivation of V-JEPA-style models~\cite{assran2025v,miao2026jepa}: a world model does not need to reconstruct future pixels in order to be useful; it can instead predict future states in a learned representation space. 
SEER-World uses the same latent prediction principle to organize multi-robot, multi-view observations into a reusable shared environmental context. For robot $i$ at time $t$, SEER Retrieval returns a context sequence
$
    \mathcal{C}_t^i
    =
    \{c_{t,1}^i,\ldots,c_{t,K}^i\},
$
where each element $c_{t,k}^i$ is an egocentric observation selected from the shared SEER Pool. 
The sequence may contain observations from robot $i$ itself and from other robots that observed related regions of the same environment. 
We also provide the corresponding action condition of the current robot,
$
    \mathbf{a}_{t,1:K-1}^i
    =
    \{a_{t,1}^i,\ldots,a_{t,K-1}^i\},
$
which tells the predictor how the current robot moves through the retrieved context.

SEER-World consists of an encoder $E_\theta$, a stop-gradient target encoder $\bar{E}_\theta$, and an action-conditioned predictor $P_\phi$. 
Given the retrieved context sequence $\mathcal{C}_t^i$, SEER-World encodes each observation, computes its stop-gradient target, and predicts the next latent state conditioned on previous latents and robot actions:
\begin{equation}
    \mathbf{z}_{t,k}^i
    = E_\theta(c_{t,k}^i), \quad
    \mathbf{y}_{t,k+1}^i
    = \mathrm{sg}\!\left(\bar{E}_\theta(c_{t,k+1}^i)\right),\quad
    \hat{\mathbf{z}}_{t,k+1}^{i}
    =P_\phi
    \left(
        \mathbf{z}_{t,1:k}^i,
        \mathbf{a}_{t,1:k}^i
    \right),
\end{equation}
$\mathrm{sg}(\cdot)$ denotes stop-gradient.
SEER-World is trained in latent space with a teacher-forcing objective, 
the loss~$\mathcal{L}_{\mathrm{tf}}$ supervises one-step latent transitions conditioned on robot actions. The overall objective combines teacher-forcing supervision with a short-horizon rollout regularization:
\begin{equation}
    \mathcal{L}_{\mathrm{world}}
    =
    \mathcal{L}_{\mathrm{tf}}
    +
    \lambda_{\mathrm{roll}}
    \mathcal{L}_{\mathrm{roll}} ,\quad
        \mathcal{L}_{\mathrm{tf}}
    =
    \frac{1}{K-1}
    \sum_{k=1}^{K-1}
    \left\|
        \hat{\mathbf{z}}_{t,k+1}^{i}
        -
        \mathbf{y}_{t,k+1}^{i}
    \right\|_1 .
\end{equation}
where the rollout loss~$\mathcal{L}_{\mathrm{roll}}$ is detailed in Appendix~\ref{app:rollout-consistency}.
After training, SEER-World outputs a shared environmental context $\mathbf{z}_{t}^{\mathrm{SEU},i}$.

\subsection{Integration with Downstream Navigation}

SEER is designed as an environmental understanding layer. After SEER-World learns a shared latent world representation~$\mathbf{z}_{t}^{\mathrm{SEU},i}$ from multi-robot observations, we use it to condition downstream navigation agents without replacing their original decision process.
This design allows each robot to retain its own policy while receiving additional shared environmental context.

Let $h_t^i$ denote the local navigation state of robot $i$, including its current observation, visual-language history, and model-specific memory.
A downstream navigation agent predicts actions from its local state and instruction $a_t^i \sim \pi_i(a \mid h_t^i, x^i)$, while our SEER-augmented agent additionally conditions on SEER tokens as $a_t^i \sim \pi_i(a \mid h_t^i, x^i, \mathbf{U}_t^i)$.~$\mathbf{U}_t^i$ denotes the SEER conditioning tokens provided to robot $i$.
SEER therefore changes what the agent knows about the shared environment, but does not change the agent's action space, controller, or decision interface.

In our implementation, $\mathbf{U}_t^i$ is constructed by querying the frozen SEER-World model with the retrieved multi-robot context and converting the resulting latent predictions into continuous soft tokens.
These tokens are inserted immediately before downstream action prediction, acting as world-conditioning prompts.
We use the same interface for both Qwen 3.5 4B~\cite{yang2025qwen3} and StreamVLN~\cite{wei2025streamvln}, demonstrating that SEER can serve as a reusable upstream representation for different types of navigation agents.
Given the retrieved context $\mathcal{C}_t^i$, SEER-World predicts one latent world hint for each candidate motion:
\begin{equation}
    \hat{\mathbf{z}}_{t+1}^{\mathrm{SEU},i}
    =
    F_{\theta,\phi}
    \left(
        \mathcal{C}_t^i,
        \mathbf{a}_{t,1:|\mathcal{C}_t^i|-1}^{i},
        a
    \right),
    a\in\mathcal{A}_{\mathrm{move}},\quad
    \mathbf{U}_t^i
    =
    G_\psi
    \left(
        \left[
            \hat{\mathbf{z}}_{t+1}^{\mathrm{SEU},i}
        \right]_{a\in\mathcal{A}_{\mathrm{move}}}
    \right),
\end{equation}
where
$
    \mathcal{A}_{\mathrm{move}}
    =
    \{\textsc{Forward},\textsc{Left},\textsc{Right}\}
$and $G_\psi$ is a lightweight projector that maps SEER latents into the hidden dimension of the downstream agent. 
In practice, the projector pools latent tokens, applies a small MLP, and adds action identity embeddings so that the downstream model can distinguish the world hints for \textsc{Forward}, \textsc{Left}, and \textsc{Right}. 
These tokens are inserted as continuous embeddings immediately before the downstream model predicts the next action, acting as a world-conditioning prompt. Thus, SEER introduces collaboration at the representation level rather than the decision level:
robots share environmental understanding, while each downstream navigation agent independently decides how to act. Detailed construction of SEER tokens and model-specific insertion strategies are provided in Appendix~\ref{app:downstream-integration-details}.

\subsection{Inverse Dynamics Analysis}
\begin{wrapfigure}{r}{.23\textwidth}
    \centering
    \captionsetup{font=footnotesize,skip=4pt}
    \includegraphics[width=\linewidth]{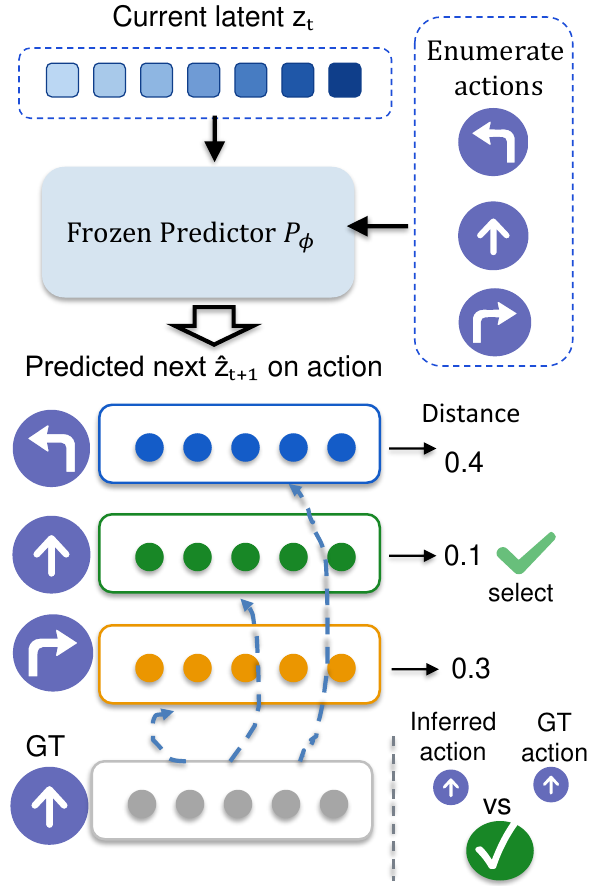}
    \caption{LIDA Illustration}
    \label{fig:ida}
\end{wrapfigure}
To assess whether the representations learned by SEER capture actionable environmental structure, we evaluate them via an inverse dynamics objective. Building on the shared latent representations produced by SEER-World with SEER Pool and Retrieval, this analysis probes whether the latent states encode sufficient information to recover agent actions. As illustrated in Figure~\ref{fig:ida}, given latent representations, an inverse dynamics model predicts the action responsible for the transition between them.
We enumerate all candidate motion actions, use the frozen SEER-World predictor to predict the next latent under each action, and select the action whose predicted latent is closest to the true next latent:
\begin{equation}
    \hat{a}_t^i
    =
    \arg\min_{a\in\mathcal{A}_{\mathrm{move}}}
    d
    \left(
        F_{\theta,\phi}(\mathbf{z}_t^i,a,\mathcal{C}_t^i),
        \mathbf{z}_{t+1}^i
    \right),
\end{equation}
where $F_{\theta,\phi}$ denotes the learned SEER-World forward latent dynamics, $\mathcal{C}_t^i$ is the retrieved context, and $d(\cdot,\cdot)$ is the $\ell_1$ distance.
We report the action recovery accuracy, denoted as Latent Inverse Dynamics Accuracy (LIDA):
\begin{equation}
    \mathrm{LIDA}
    =
    \frac{\sum_{(\mathbf{z}_t,a_t,\mathbf{z}_{t+1})\in\mathcal{D}_{\mathrm{eval}}}
    \mathbf{1}[\hat{a}_t=a_t]}{|\mathcal{D}_{\mathrm{eval}}|}
    .
\end{equation}
A higher LIDA indicates that different actions induce distinguishable transitions in latent space, suggesting that the learned shared environmental representations capture navigation-relevant dynamics rather than merely appearance-level semantics. 
Additional details are provided in Appendix~\ref{app:inverse_dynamics}.

%% file: sec/4_experiments.tex
\section{Experiments}
\subsection{Experimental Setup}
\input{tables/table1_vlnce}
We evaluate SEER in both simulated VLN environments and real-world robot deployments. 
The simulated evaluation contains three parts: standard single-robot VLN, multi-robot VLN constructed from paired R2R trajectories, and downstream integration with different navigation agents. Detailed construction of multi-robot episodes  are provided in Appendix~\ref{app:multi_robot_episode_construction}.
For standard VLN, we follow the continuous VLN-CE setting and report results on the Val-Unseen splits of R2R-CE~\citep{anderson2018vision,krantz2020beyond} and RxR-CE~\citep{ku2020rxr} in the Habitat simulator~\citep{savva2019habitat}. 
We use standard VLN metrics, including Navigation Error (NE), Success Rate (SR), Oracle Success Rate (OSR), Success weighted by Path Length (SPL), and normalized Dynamic Time Warping (nDTW)~\citep{anderson2018vision,ilharco2019dtw}. 
These single-robot evaluations test whether adding SEER preserves the original navigation ability of downstream agents.

For multi-robot evaluation, we use the paired R2R episodes constructed by ourself. 
The evaluation set is divided into four subsets according to the spatial relationship between the two robots: \emph{different-scene}, \emph{same-scene non-intersecting}, \emph{same-scene intersecting}, and \emph{same/sub-trajectory}. 
These subsets allow us to test whether shared environmental understanding helps under different levels of shared context, from no shared scene to highly overlapping trajectories. 
Following the spirit of collaborative navigation evaluation~\citep{wang2026conavbench}, we report both individual and collaborative success. 
For a paired episode $e$, let $s_e^A,s_e^B\in\{0,1\}$ denote whether robot A and robot B successfully reach their respective goals. 
We define Individual Success Rate (ISR) and Collaborative Success Rate (CSR) as
\begin{equation}
    \mathrm{ISR}
    =
    \frac{1}{2|\mathcal{D}|}
    \sum_{e\in\mathcal{D}}
    (s_e^A+s_e^B),
    \qquad
    \mathrm{CSR}
    =
    \frac{1}{|\mathcal{D}|}
    \sum_{e\in\mathcal{D}}
    s_e^A s_e^B .
\end{equation}
ISR measures whether each robot can still complete its own task, while CSR measures whether both robots succeed in the paired episode. We emphasize that our paired-R2R evaluation is not intended to model full physical multi-robot coordination; instead, it provides a controlled abstraction for isolating SEER’s core question: whether independently collected egocentric observations can be transformed into reusable shared environmental context for decentralized navigation.

We instantiate downstream navigation with two agents: a specialized StreamVLN agent~\citep{wei2025streamvln} and a general Qwen3.5-based agent~\citep{yang2025qwen3}, testing whether SEER is reusable rather than model-specific. We further conduct real-world indoor experiments on two embodiments, AgileX LiMO Pro and Unitree Go2, which performs inference and returns actions to robot-specific low-level controllers via ROS~2. Details on splits, evaluation, deployment, and implementation are in Appendix~\ref{app:experimental_setup}.

\input{tables/table2_multi}
\input{tables/table3_lida}
\subsection{Task-Level Validation: Shared Environmental Understanding for Navigation}
\textbf{Single-agent preservation.}
A key requirement of SEER is that it should not degrade ordinary single-agent navigation. 
In this sense, single-agent navigation can be viewed as a special case of our setting, where the robot mainly relies on its own observation context and receives no useful cross-robot information.
Table~\ref{tab:comp-vlnce} evaluates this property, 
when SEER is added to StreamVLN, the resulting agent preserves the strong performance of the original navigation model and improves the main success-oriented metrics. 
These results show that SEER tokens provide useful environmental context while preserving the downstream action space and output format. SEER-World is trained independently of downstream navigation policies and frozen during downstream adaptation, while the projector and downstream navigation agent are optimized. Here, policy-agnosticism refers to reusing the same shared world model across downstream navigation policies. 
\textbf{Multi-robot navigation improvement.}
Table~\ref{tab:multi_agent_performance} evaluates SEER in paired multi-robot episodes under varying shared environmental context. Across all subsets, SEER improves both ISR and CSR. ISR gains show that each robot benefits from shared environmental context while solving its own task, whereas CSR gains indicate that both robots are more likely to succeed in the same paired episode. SEER achieves stronger gains in same-scene and overlapped-trajectory settings, where cross-robot observations are more relevant.

\textbf{Compatibility with different downstream agents.}
We evaluate SEER with both a specialized VLN agent (StreamVLN) and a general VLM-based navigation agent (Qwen 3.5), using the same integration strategy. Results show that SEER improves StreamVLN and brings even larger gains to the Qwen 3.5 agent. This supports that SEER is not a model-specific trick, but a reusable environmental understanding layer for different downstream navigation agents. It enables collaborative navigation by sharing latent world understanding, without sharing actions, histories, or a centralized policy.

\subsection{Representation-Level Validation: Action-Relevant Latent Dynamics}
We introduce a representation-level inverse dynamics probe to test whether SEER encodes action-relevant environmental structure. Since the latent state is not directly interpretable, we freeze the learned forward latent dynamics and, for each transition, enumerate candidate actions. The action whose predicted next latent is closest to the true next latent is selected. High recovery accuracy therefore indicates that the latent space preserves how different navigation actions transform the world state, rather than merely encoding static appearance.

Table~\ref{tab:inverse_dynamics_analysis} shows that the probe is non-trivial: V-JEPA2 achieves reasonable accuracy, indicating that generic predictive video features capture motion cues, but still lags behind SEER. \emph{Ours w/o Retrieval} ( replaced by uniform random sampling ) already performs well, suggesting that SEER-World learns discriminative latent dynamics via multi-view prediction. SEER Retrieval further improves both single- and dual-agent settings, isolating the benefit of shared retrieval: relevant observations from the anonymous SEER Pool provide cleaner cross-robot context than the world model alone. Importantly, LIDA is a held-out counterfactual action recovery test across diverse spatial configurations, not a direct training-loss measure, making high scores unlikely to stem only from action bias, visual repetition, or retrieval leakage.

\input{tables/table4_other_share}
\begin{figure}[t!]
    \centering
    \includegraphics[width=1\textwidth]{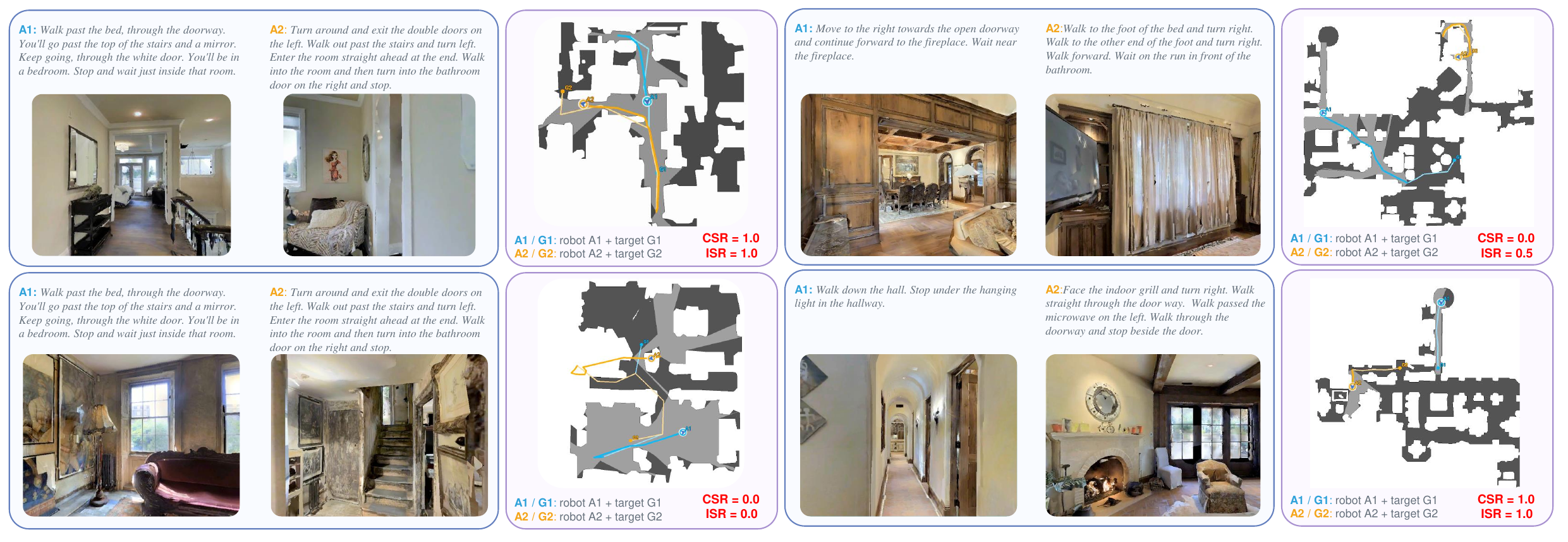}
\caption{\small{Qualitative Analysis of Success and Failure Cases on different Dual-Agent State.}}
\label{fig:vis}
\end{figure}
\subsection{What Should Robots Share: World Understanding vs. Agent-Centric Traces}
Table~\ref{tab:share_what} compares SEER with several sharing strategies. We position these baselines as controlled comparisons against representative agent-centric sharing strategies, while leaving stronger learned communication, map-sharing, and centralized multi-agent architectures as complementary directions for future work rather than direct targets of this representation-layer study. Sharing actions only reveals what another robot did, without explaining the environmental structure that made the action useful. Sharing spatial structure$^*$ improves over these simple signals by providing explicit geometric context, but it requires an additional 3D expert and mainly captures static layout, without modeling how actions change the world state.
SEER showing that collaboration benefits more from shared world understanding than from sharing raw histories, actions, or hand-crafted maps. Notably, semantic or topological maps are not mutually exclusive with SEER; they can be viewed as complementary external structures, while SEER provides a learned, action-aware representation layer.

%% file: tables/table1_vlnce.tex
\begin{table*}[!t]
\centering
\scriptsize
\setlength{\tabcolsep}{1.25pt}
\renewcommand{\arraystretch}{1.15}
\caption{Comparison on VLN-CE R2R and RxR \textbf{Val-Unseen} splits. Pano, Odo and D respectively represent panoramic view, odometry and depth, SRGB denotes monocular RGB.}
\resizebox{0.88\textwidth}{!}{%
 \begin{tabular}{l|cccc|cccc|cccc}
\toprule
\multirow{2}{*}{Method} &
\multicolumn{4}{c|}{Observation} &
\multicolumn{4}{c|}{R2R Val-Unseen} &
\multicolumn{4}{c}{RxR Val-Unseen} \\
\cmidrule(lr){2-5}\cmidrule(lr){6-9}\cmidrule(lr){10-13}
& Pano. & Odo. & D. & SRGB
& NE$\downarrow$ & OS$\uparrow$ & SR$\uparrow$ & SPL$\uparrow$
& NE$\downarrow$ & SR$\uparrow$ & SPL$\uparrow$ & nDTW$\uparrow$ \\
\midrule

CMA$ $~\cite{hong2022bridging}
& $\checkmark$ & $\checkmark$ & $\checkmark$ &
& 6.20 & 52.0 & 41.0 & 36.0
& 8.76 & 26.5 & 22.1 & 47.0 \\

VLN-BERT$ $~\cite{hong2022bridging}
& $\checkmark$ & $\checkmark$ & $\checkmark$ &
& 5.74 & 53.0 & 44.0 & 39.0
& 8.98 & 27.0 & 22.6 & 46.7 \\

Sim2Sim$ $~\cite{krantz2022sim2sim}
& $\checkmark$ & $\checkmark$ & $\checkmark$ &
& 6.07 & 52.0 & 43.0 & 36.0
& -- & -- & -- & -- \\

$\text{Ego}^2\text{-Map}$~\cite{hong2023ego2map}
& $\checkmark$ & $\checkmark$ & $\checkmark$ &
& 5.54 & 56.0 & 47.0 & 41.0
& -- & -- & -- & -- \\

DreamWalker~\cite{wang2023dreamwalker}
& $\checkmark$ & $\checkmark$ & $\checkmark$ &
& 5.53 & 59.0 & 49.0 & 44.0
& -- & -- & -- & -- \\

ETPNav$ $~\cite{an2023etpnav}
& $\checkmark$ & $\checkmark$ & $\checkmark$ &
& 4.71 & 65.0 & 57.0 & 49.0
& 5.64 & 54.7 & 44.8 & 61.9 \\

ScaleVLN$ $~\cite{wang2023scalevln}
& $\checkmark$ & $\checkmark$ & $\checkmark$ &
& 4.80 & -- & 55.0 & 51.0
& -- & -- & -- & -- \\

\midrule

InstructNav~\cite{long2024instructnav}
& $\checkmark$ & $\checkmark$ & $\checkmark$ & $\checkmark$
& 6.89 & -- & 31.0 & 24.0
& -- & -- & -- & -- \\


LAW~\cite{raychaudhuri2021law}
&  & $\checkmark$ & $\checkmark$ & $\checkmark$
& 6.83 & 44.0 & 35.0 & 31.0
& 10.90 & 8.0 & 8.0 & 38.0 \\



Sim2Real~\cite{wang2024sim2real3dff}
&  & $\checkmark$ & $\checkmark$ & $\checkmark$
& 5.95 & 55.8 & 44.9 & 30.4
& 8.79 & 25.5 & 18.1 & -- \\

Seq2Seq~\cite{krantz2020beyond}
&  &  & $\checkmark$ & $\checkmark$
& 7.77 & 37.0 & 25.0 & 22.0
& 12.10 & 13.9 & 11.9 & 30.8 \\

CMA~\cite{krantz2020beyond}
&  &  & $\checkmark$ & $\checkmark$
& 7.37 & 40.0 & 32.0 & 30.0
& -- & -- & -- & -- \\

\midrule

NaVid~\cite{zhang2024navid}
&  &  &  & $\checkmark$
& 5.47 & 49.1 & 37.4 & 35.9
& -- & -- & -- & -- \\

AO-Planner~\cite{chen2025aoplanner}
& $\checkmark$ &  & $\checkmark$ &
& 5.55 & 59.0 & 47.0 & 33.0
& 7.06 & 43.3 & 30.5 & 50.1 \\


MapNav~\cite{zhang2025mapnav}
&  &  &  & $\checkmark$
& 4.93 & 53.0 & 39.7 & 37.2
& -- & -- & -- & -- \\

NaVid-4D~\cite{liu2025navid4d}
&  &  & $\checkmark$ & $\checkmark$
& 5.99 & 55.7 & 43.8 & 37.1
& -- & -- & -- & -- \\

NavMorph~\cite{yao2025navmorph}
&  &  & $\checkmark$ & $\checkmark$
& 5.75 & 56.9 & 47.9 & 33.2
& 8.85 & 30.8 & 22.8 & 44.2 \\

Uni-NaVid~\cite{zhang2025uninavid}
&  &  &  & $\checkmark$
& 5.58 & 53.3 & 47.0 & 42.7
& 6.24 & 48.7 & 40.9 & -- \\

NaVILA~\cite{cheng2025navila}
&  &  &  & $\checkmark$
& 5.22 & 62.5 & 54.0 & 49.0
& 6.77 & 49.3 & 44.0 & 58.8 \\
\midrule
StreamVLN 7B~\cite{wei2025streamvln}
&  &  &  & $\checkmark$
& 4.98 & 64.2 & 56.9 & 51.9
& 6.22 & 52.9 & 46.0 & 61.9 \\

\rowcolor{LightGray}

+SEER(ours)
&  &  &  & $\checkmark$
& 4.89&65.6&58.1&52.5&6.17&54.3&47.2&62.1 \\
\midrule
Qwen3.5 4B
&  &  &  & $\checkmark$
& 7.35&49.8&40.3&35.9&8.01&39.7&32.5&45.6 \\
\rowcolor{LightGray}

+SEER(ours)
&  &  &  & $\checkmark$
& 6.97&53.2&44.9&39.8&7.53&42.4&35.6&49.4\\
\bottomrule
\end{tabular}}
\label{tab:comp-vlnce}
\end{table*}

%% file: tables/table2_multi.tex
\begin{table*}[!t]
\centering
\footnotesize
\setlength{\tabcolsep}{2.0pt}
\renewcommand{\arraystretch}{1.15}
\caption{\small{Multi-agent performance comparison. For dual-agent state tests, CSR and ISR denote complete success rate and individual success rate, respectively. SEER significantly boosts multi-agent navigation performance across all dual-agent settings.} }
\resizebox{\textwidth}{!}{%
\begin{tabular}{l|cc|cc|cc|cc|cc}
\toprule
\multirow{3}{*}{Method} &
\multicolumn{2}{c|}{Single-Agent} &
\multicolumn{8}{c}{Dual-Agent State Test} \\
\cmidrule(lr){2-3}\cmidrule(lr){4-11}
&
\multicolumn{1}{c}{R2R Unseen} &
\multicolumn{1}{c|}{RxR Unseen} &
\multicolumn{2}{c|}{Different Scenes} &
\multicolumn{2}{c|}{Same Scene w/o Intersection} &
\multicolumn{2}{c|}{Same Scene w/ Intersection} &
\multicolumn{2}{c}{Overlapped Trajectories} \\
\cmidrule(lr){2-3}\cmidrule(lr){4-5}\cmidrule(lr){6-7}\cmidrule(lr){8-9}\cmidrule(lr){10-11}
&
SR$\uparrow$ &
SR$\uparrow$ &
CSR$\uparrow$ & ISR$\uparrow$ &
CSR$\uparrow$ & ISR$\uparrow$ &
CSR$\uparrow$ & ISR$\uparrow$ &
CSR$\uparrow$ & ISR$\uparrow$ \\
\midrule

Random&2.50&5.89&0.0&3.05&0.0&2.15&0.0&1.60&0.0&1.40\\

StreamVLN
& 56.9
& 52.9
& 28.8
& 52.2
& 28.2
& 51.2
& 31.2
& 55.7
& 30.3
& 54.9 \\

+SEER 
& 58.1
& 54.3
& 32.9 {\scriptsize (+4.1)}
& 56.1 {\scriptsize (+3.9)}
& 32.6 {\scriptsize (+4.4)}
& 55.8 {\scriptsize (+4.6)}
& 35.9 {\scriptsize (+4.7)}
& 60.7 {\scriptsize (+5.0)}
& 35.4 {\scriptsize (+5.1)}
& 60.1 {\scriptsize (+5.2)} \\

\midrule

Qwen3.5-4B
& 40.3
& 39.7
& 17.0
& 39.7
& 17.7
& 40.6
& 18.1
& 41.4
& 17.2
& 40.1 \\

+SEER 
& 44.9
& 42.4
& 21.6 {\scriptsize (+4.6)}
& 43.8 {\scriptsize (+4.1)}
& 22.1 {\scriptsize (+4.4)}
& 45.3 {\scriptsize (+4.7)}
& 23.9 {\scriptsize (+5.8)}
& 47.5 {\scriptsize (+6.1)}
& 22.4 {\scriptsize (+5.2)}
& 45.6 {\scriptsize (+5.5)} \\

\bottomrule
\end{tabular}%
}
\label{tab:multi_agent_performance}
\end{table*}

%% file: tables/table3_lida.tex
\begin{table*}[!t]
\centering
\footnotesize
\setlength{\tabcolsep}{2.0pt}
\renewcommand{\arraystretch}{1.15}
\caption{\small{Inverse dynamics analysis.SEER-World dramatically improves inverse dynamics accuracy by capturing how actions transform the environment in latent space. }}
\resizebox{\textwidth}{!}{%
\begin{tabular}{l|cc|cc|cc|cc|cc}
\toprule
\multirow{3}{*}{Method} &
\multicolumn{2}{c|}{Single-Agent} &
\multicolumn{8}{c}{Dual-Agent State Test} \\
\cmidrule(lr){2-3}\cmidrule(lr){4-11}
&
\multicolumn{1}{c}{R2R Unseen} &
\multicolumn{1}{c|}{RxR Unseen} &
\multicolumn{2}{c|}{Different Scenes} &
\multicolumn{2}{c|}{Same Scene w/o Intersection} &
\multicolumn{2}{c|}{Same Scene w/ Intersection} &
\multicolumn{2}{c}{Overlapped  Trajectories} \\
\cmidrule(lr){2-3}\cmidrule(lr){4-5}\cmidrule(lr){6-7}\cmidrule(lr){8-9}\cmidrule(lr){10-11}
&
SR$\uparrow$ &
SR$\uparrow$ &
CSR$\uparrow$ & ISR$\uparrow$ &
CSR$\uparrow$ & ISR$\uparrow$ &
CSR$\uparrow$ & ISR$\uparrow$ &
CSR$\uparrow$ & ISR$\uparrow$ \\
\midrule

Random
& 33.8
& 33.4
& 11.3
& 33.6
& 11.0
& 33.2
& 11.4
& 33.9
& 11.2
& 33.4 \\

V-JEPA2
& 78.2
& 77.2
& 66.8
& 77.4
& 68.3
& 78.7
& 68.0
& 78.4
& 70.7
& 79.5 \\

Ours w/o SEER Retrieval
& 92.1
& 91.9
& 89.1
& 91.5
& 89.6
& 91.8
& 88.8
& 91.3
& 90.1
& 91.9 \\

\textbf{Ours}
& \textbf{95.4}
& \textbf{94.7}
& \textbf{94.9}
& \textbf{95.4}
& \textbf{94.8}
& \textbf{95.3}
& \textbf{95.9}
& \textbf{95.9}
& \textbf{96.0}
& \textbf{96.0} \\

\bottomrule
\end{tabular}%
}
\label{tab:inverse_dynamics_analysis}
\end{table*}

%% file: tables/table4_other_share.tex
\begin{table*}[!t]
\centering
\footnotesize
\setlength{\tabcolsep}{2.0pt}
\renewcommand{\arraystretch}{1.15}
\caption{\small{Comparison of different sharing strategies on Qwen3.5-4B. $^*$Share Spatial Structure uses an additional 3D expert InternVL3 8B ~\citep{zhu2025internvl3} to extract and share spatial descriptions from observations.}}
\resizebox{0.9\textwidth}{!}{%
\begin{tabular}{l|cc|cc|cc|cc}
\toprule
\multirow{3}{*}{Method} &
\multicolumn{8}{c}{Dual-Agent State Test} \\
\cmidrule(lr){2-9}
&
\multicolumn{2}{c|}{Different Scenes} &
\multicolumn{2}{c|}{Same Scene w/o Intersection} &
\multicolumn{2}{c|}{Same Scene w/ Intersection} &
\multicolumn{2}{c}{Overlapped Trajectories} \\
\cmidrule(lr){2-3}\cmidrule(lr){4-5}\cmidrule(lr){6-7}\cmidrule(lr){8-9}
&
CSR$\uparrow$ & ISR$\uparrow$ &
CSR$\uparrow$ & ISR$\uparrow$ &
CSR$\uparrow$ & ISR$\uparrow$ &
CSR$\uparrow$ & ISR$\uparrow$ \\
\midrule

No Sharing
& 17.0
& 39.7
& 17.7
& 40.6
& 18.1
& 41.4
& 17.2
& 40.1  \\

Share History
& 16.2 & 38.9 & 17.3 & 39.5 & 17.8 & 40.9 &17.4  & 40.3 \\

Share Action 
& 17.6 & 40.0 & 18.1 & 40.4 & 19.4 & 42.2 & 19.7 & 42.4 \\ 
Share Observation
& 18.4 & 40.5 & 19.2 & 41.1 & 19.8 & 43.4 & 20.3 & 43.4 \\
$^*$Share Spatial
& 19.6 & 41.4 & 20.7 & 42.8 & 20.6 & 45.1 & 20.9 & 44.7 \\
\rowcolor{LightGray}
SEER

& 21.6 & 43.8& 22.1 & 45.3 & 23.9& 47.5 & 22.4& 45.6 \\

\bottomrule
\end{tabular}%
}
\label{tab:share_what}
\end{table*}

%% file: sec/appendix.tex
\clearpage
\section{Appendix}
\label{app:supplementary}

This appendix details SEER-World training, downstream integration, inverse dynamics evaluation, paired-episode construction, and physical deployment. The notation follows the main text.

\subsection{Rollout Consistency Regularization}
\label{app:rollout-consistency}

\textbf{Short-horizon prediction.}
The teacher-forcing objective supervises one-step latent transitions using observed states as predictor inputs. We additionally regularize short autoregressive rollouts, in which predicted latents are fed back to the predictor. Starting from $\mathbf{z}_t^i$ and conditioning on the action sequence $\mathbf{a}_{t:t+R-1}^i$, the $R$-step prediction and its consistency loss are
\begin{equation}
    \widetilde{\mathbf{z}}_{t+R}^i
    = \mathrm{Roll}_R\!\left(P_\phi;\mathbf{z}_t^i,
        \mathbf{a}_{t:t+R-1}^i\right),
    \qquad
    \mathcal{L}_{\mathrm{roll}}
    = \left\|\widetilde{\mathbf{z}}_{t+R}^i
        -\mathbf{y}_{t+R}^i\right\|_1,
\end{equation}
where $\mathrm{Roll}_R$ applies the action-conditioned predictor recursively and $\mathbf{y}_{t+R}^i$ is the stop-gradient target latent. This term encourages consistency over successive predicted transitions.

\textbf{Training objective.}
SEER-World combines the teacher-forcing loss defined in the main text with rollout consistency:
\begin{equation}
    \mathcal{L}_{\mathrm{world}}
    = \mathcal{L}_{\mathrm{tf}}
    + \lambda_{\mathrm{roll}}\mathcal{L}_{\mathrm{roll}}.
\end{equation}
The resulting encoder and predictor form $F_{\theta,\phi}$, which converts retrieved observations and action conditions into latent environmental context for downstream navigation.

\subsection{Downstream Navigation Integration}
\label{app:downstream-integration-details}

\textbf{Candidate-action predictions.}
For robot $i$, the retrieved context $\mathcal{C}_t^i$ and its action condition are passed to the frozen SEER-World model. We query the candidate motions $\mathcal{A}_{\mathrm{move}}=\{\textsc{Forward},\textsc{Left},\textsc{Right}\}$ to obtain action-conditioned world hints:
\begin{equation}
    \widehat{\mathbf{z}}_{t+1}^{i,a}
    = F_{\theta,\phi}\!\left(
        \mathcal{C}_t^i,
        \mathbf{a}_{t,1:|\mathcal{C}_t^i|-1}^i,a\right),
    \qquad a\in\mathcal{A}_{\mathrm{move}}.
\end{equation}
The downstream policy also predicts $\textsc{Stop}$ as a termination action; latent motion hints are constructed for the three movement actions.

\textbf{Soft-token construction.}
The projector pools the predicted latent tokens, applies a small MLP, and adds action identity embeddings. It maps the resulting hints into the hidden dimension of the downstream navigation model:
\begin{equation}
    \mathbf{U}_t^i
    = G_\psi\!\left(
        \left[\widehat{\mathbf{z}}_{t+1}^{i,a}\right]_
        {a\in\mathcal{A}_{\mathrm{move}}}\right).
\end{equation}
The action embeddings distinguish the predicted effects of moving forward, turning left, and turning right.

\textbf{Training and insertion.}
During downstream training, SEER-World is frozen, while the projector and downstream navigation agent are optimized. Thus, the same independently trained environmental model can be reused across downstream policies, with model-specific adaptation at the token interface. For Qwen3.5-4B, SEER tokens are inserted immediately before the assistant's action response. For StreamVLN, they are inserted at each action-prediction round. Both agents retain their original action spaces and output formats, and each robot selects actions from its own local state, instruction, and SEER tokens.

\subsection{Inverse Dynamics Evaluation}
\label{app:inverse_dynamics}

\textbf{Action recovery.}
Latent Inverse Dynamics Accuracy (LIDA) measures whether the learned forward model distinguishes the environmental effects of different navigation actions. For each evaluation transition, we encode the current and next observations as $\mathbf{z}_t^i=E_\theta(o_t^i)$ and $\mathbf{z}_{t+1}^i=E_\theta(o_{t+1}^i)$. We then enumerate the candidate motions using the frozen predictor and recover the action whose predicted latent is closest to the observed next latent:
\begin{equation}
    \widetilde{\mathbf{z}}_{t+1}^{i,a}
    = F_{\theta,\phi}(\mathbf{z}_t^i,a,\mathcal{C}_t^i),
    \qquad
    \widehat{a}_t^i
    = \arg\min_{a\in\mathcal{A}_{\mathrm{move}}}
        \left\|\widetilde{\mathbf{z}}_{t+1}^{i,a}
            -\mathbf{z}_{t+1}^i\right\|_1.
\end{equation}
This evaluation uses the $\ell_1$ distance, consistent with the main text. Action recovery is computed directly from the frozen forward model by enumerating candidate motions. The next observation provides the evaluation target; retrieved context is formed from observations available at time $t$ under the protocol in Appendix~\ref{app:multi_robot_episode_construction}.

\textbf{Accuracy.}
For an evaluation set $\mathcal{D}_{\mathrm{eval}}$ of latent transitions, LIDA is
\begin{equation}
    \mathrm{LIDA}
    = \frac{1}{|\mathcal{D}_{\mathrm{eval}}|}
      \sum_{(\mathbf{z}_t,a_t,\mathbf{z}_{t+1})\in\mathcal{D}_{\mathrm{eval}}}
      \mathbf{1}[\widehat{a}_t=a_t].
\end{equation}
Higher accuracy indicates that the predicted latent transitions preserve action-relevant environmental structure. We apply this evaluation to the single-agent data and the four paired-agent configurations described below.

\subsection{Paired Multi-Robot Episode Construction}
\label{app:multi_robot_episode_construction}

\textbf{Episode definition.}
Each paired episode combines two original R2R episodes, assigned to robots A and B. Each robot retains its own instruction, start state, goal, and reference trajectory. The reference paths determine the pairing category, while shared observations are accumulated during execution. The same construction supports inverse dynamics analysis and downstream navigation evaluation.

\textbf{Path representation.}
For geometric analysis, we concatenate the start position, reference path, and goal positions, then remove consecutive duplicate points:
\begin{equation}
    \rho = \mathrm{Dedup}\!\left(
        [p_{\mathrm{start}};\rho_{\mathrm{ref}};p_{\mathrm{goal}}]
    \right).
\end{equation}
Including the endpoints preserves intersections and overlaps at the start or goal. The resulting effective paths are used consistently for pair filtering and visualization.

\textbf{Training and evaluation splits.}
Training pairs are sampled within scenes from the R2R train split. Duplicate pairs are removed with A/B swaps treated as equivalent, and identical or reversed-identical paths are filtered out. Evaluation pairs are constructed from R2R val-unseen and grouped into four configurations:

\textbf{Different scenes.}
The robots navigate in different scenes, $\mathrm{scene}^A\neq\mathrm{scene}^B$. This configuration measures navigation when the shared pool contains observations from unrelated environments.

\textbf{Same scene without intersection.}
The robots share a scene, but their effective paths do not intersect. This configuration provides scene-level context with spatially separated routes.

\textbf{Same scene with intersection.}
The robots share a scene and their effective paths intersect. This configuration provides opportunities to reuse observations of common local regions.

\textbf{Overlapped trajectories.}
The effective paths are identical, reversed-identical, or related by a continuous sub-trajectory. Paired episodes have different instruction texts. This configuration corresponds to the same/sub-trajectory setting and provides highly overlapping environmental context.

\textbf{Geometric checks and deduplication.}
Path intersection is computed between polyline segments on the horizontal XZ plane. A projected intersection is accepted only when the corresponding vertical separation satisfies $|y^A-y^B|\leq\epsilon_{\mathrm{floor}}$, which separates crossings on different floors. We remove duplicate episode pairs and repeated geometric path pairs; identical and reversed-identical paths are canonicalized and filtered when they are outside the target category. Sampling uses a fixed random seed, and path coordinates are rounded before canonicalization to reduce floating-point instability. Pair construction preserves the R2R train/val-unseen split.

\textbf{Causal shared memory.}
At decision step $t$, the bounded SEER Pool contains observations already uploaded by either robot, with timestamps no later than $t$:
\begin{equation}
    \mathcal{M}_t
    \subseteq
    \{(o_s^j,s)\mid j\in\{A,B\},\ s\leq t\},
    \qquad |\mathcal{M}_t|\leq 1{,}024.
\end{equation}
An observation captured at the current timestep becomes eligible for retrieval once uploaded. Each robot retrieves a bounded context from this pool and selects its own action using its local state, instruction, and SEER tokens. Algorithm~\ref{alg:causal-seer-evaluation} summarizes the synchronized evaluation protocol.

\begin{algorithm}[tbp]
\caption{Causal SEER evaluation}
\label{alg:causal-seer-evaluation}
\begin{algorithmic}[1]
\STATE Initialize shared memory $\mathcal{M}_0\leftarrow\emptyset$
\FOR{$t=1,\ldots,T$}
    \STATE $\mathcal{M}_t\leftarrow\mathcal{M}_{t-1}$
    \FOR{each robot $i$}
        \STATE Upload $(o_t^i,t)$ to $\mathcal{M}_t$, retaining the FIFO capacity
    \ENDFOR
    \FOR{each robot $i$}
        \STATE Retrieve context $\mathcal{C}_t^i$ from $\mathcal{M}_t$ using $o_t^i$
        \STATE Form $\mathbf{U}_t^i$ from SEER-World predictions and the projector
        \STATE Select $a_t^i\sim\pi_i(a\mid h_t^i,x^i,\mathbf{U}_t^i)$
        \STATE Execute $a_t^i$ with the robot-specific controller
    \ENDFOR
\ENDFOR
\end{algorithmic}
\end{algorithm}

\subsection{Experimental Setup}
\label{app:experimental_setup}

\textbf{Simulation and agents.}
Single-robot navigation is evaluated on the original R2R-CE and RxR-CE Val-Unseen episodes in Habitat~\citep{savva2019habitat}. Multi-robot navigation uses the four paired R2R val-unseen configurations in Appendix~\ref{app:multi_robot_episode_construction}. We integrate SEER with StreamVLN~\citep{wei2025streamvln} and Qwen3.5-4B~\citep{yang2025qwen3}, using the soft-token interface in Appendix~\ref{app:downstream-integration-details}.

\textbf{Retrieval settings.}
All main experiments use a FIFO pool of 1,024 observation records and a total context budget of $K=8$ observations. Retrieval uses weighted sampling without replacement, with weights renormalized over the remaining records after each draw. The temporal and spatial coefficients are $\gamma_T=\gamma_S=1$, and the weight-floor parameter is $\epsilon=0.1$. These settings are shared across both downstream agents and all four paired-agent configurations. When cross-robot observations are unavailable, recent self-history supplies the remaining context.

\textbf{Navigation metrics.}
Navigation Error (NE) is the final geodesic distance to the goal. Success Rate (SR) measures the fraction of episodes ending within 3\,m of the goal, while Oracle Success Rate (OSR) uses the closest point along the executed trajectory. Success weighted by Path Length (SPL) additionally accounts for path efficiency, and normalized Dynamic Time Warping (nDTW) measures similarity to the reference trajectory. For paired episodes, ISR averages individual goal-reaching outcomes, and CSR counts episodes in which both robots succeed, as defined in the main text.

\textbf{Sharing baselines.}
Table~\ref{tab:share_what} compares no sharing, history sharing, action sharing, observation sharing, spatial-description sharing, and SEER using Qwen3.5-4B. The spatial-description baseline uses InternVL3-8B~\citep{zhu2025internvl3} as an additional expert. These comparisons evaluate different shared inputs within the same downstream navigation interface.

\subsection{Real-World Deployment}
\label{sec:real_world}

\begin{figure}[tbp]
    \centering
    \includegraphics[width=0.94\linewidth,height=0.36\textheight,keepaspectratio]{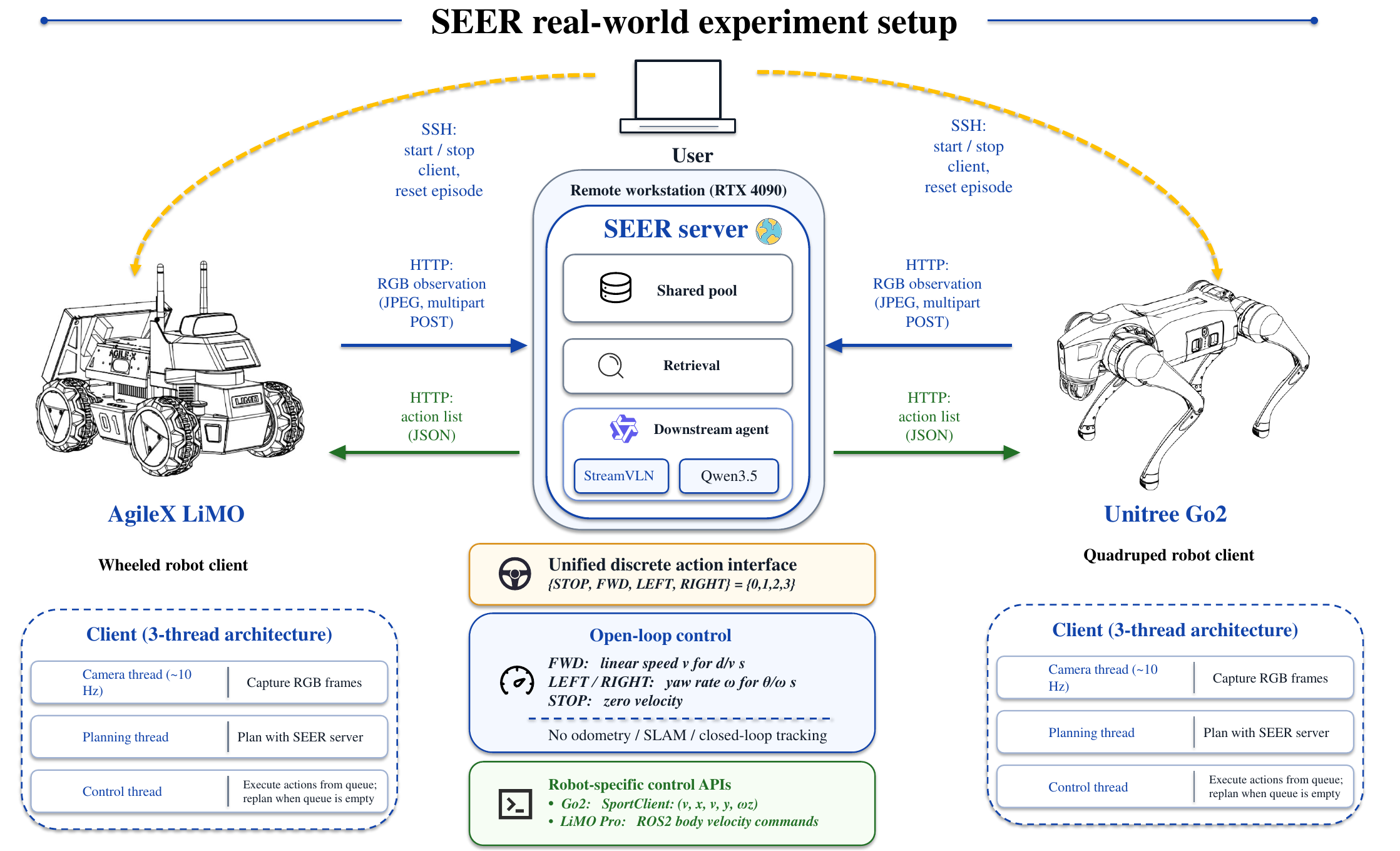}
    \caption{Real-world deployment of SEER. LiMO Pro and Go2 upload RGB observations to a shared server and execute the returned high-level actions through their platform-specific control interfaces.}
    \label{fig:real-world-setup}
    \label{fig:Real-world }
\end{figure}

\textbf{Platforms and server.}
We deploy SEER on a wheeled AgileX LiMO Pro and a quadruped Unitree Go2 in indoor corridors and rooms. LiMO Pro is equipped with LiDAR and an Intel RealSense D435 camera~\citep{agilex_limo_pro}; Go2 provides an onboard wide-angle camera. SEER consumes egocentric RGB observations from both platforms. A remote workstation with an NVIDIA RTX 4090 runs the shared pool, retrieval, SEER-World, and downstream navigation models. Robot clients handle image capture and physical action execution, as shown in Figure~\ref{fig:real-world-setup}.

\textbf{Observation and action exchange.}
At each decision step, the client sends a JPEG-compressed RGB image in a multipart HTTP POST request. The server updates the pool, retrieves context, constructs SEER tokens, and evaluates the downstream agent for that robot. Both StreamVLN and Qwen3.5-4B use the discrete action interface $\{\textsc{Forward},\textsc{Left},\textsc{Right},\textsc{Stop}\}$, encoded as $\{1,2,3,0\}$. The server returns a JSON action list, which the client executes sequentially before requesting the next prediction. SSH is used to start or stop clients and reset episodes.

\textbf{Robot-specific execution.}
High-level actions are mapped to timed velocity commands. A forward action commands linear speed $v$ for $d/v$ seconds; a turn commands yaw rate $\omega$ for $\theta/\omega$ seconds; a stop action sets commanded velocities to zero. Go2 uses Unitree's SportClient API with $(v_x,v_y,\omega_z)$ commands. Its forward settings are $v=0.3$\,m/s and $d=0.25$\,m, and turns use $\omega=0.5$\,rad/s and $\theta=15^\circ$, followed by a short settling delay. LiMO Pro uses its ROS~2 velocity-control interface with the same action-to-duration mapping and platform-specific speeds. Navigation actions are executed through open-loop velocity control without SLAM or closed-loop waypoint tracking.

\textbf{Client scheduling.}
Each client uses three threads. The camera thread updates the latest RGB frame at approximately 10\,Hz; the planning thread sends the latest frame when a new prediction is requested; and the control thread executes the action queue and triggers replanning when the queue becomes empty. This deployment keeps the navigation interface consistent across wheeled and legged platforms while retaining their respective low-level controllers.

\subsection{Additional Task-Level Analysis}
\label{app:task_level_validation}

\textbf{Single-agent navigation.}
Table~\ref{tab:comp-vlnce} reports both downstream agents with and without SEER on R2R-CE and RxR-CE. On R2R-CE, SR increases from $56.9\%$ to $58.1\%$ for StreamVLN and from $40.3\%$ to $44.9\%$ for Qwen3.5-4B. On RxR-CE, SR increases from $52.9\%$ to $54.3\%$ and from $39.7\%$ to $42.4\%$, respectively. The corresponding SPL results also improve. These results show that shared-world conditioning remains useful when the context is formed from a single robot's observations.

\textbf{Paired-agent navigation.}
Table~\ref{tab:multi_agent_performance} reports higher ISR and CSR with SEER in all four configurations. For example, in the same-scene intersecting setting, StreamVLN's CSR increases from $31.2\%$ to $35.9\%$, while Qwen3.5-4B's CSR increases from $18.1\%$ to $23.9\%$. The corresponding gains are $4.7$ and $5.8$ percentage points. Different-scene pairs provide a control with unrelated cross-robot observations, while intersecting and overlapped routes provide shared local evidence. ISR summarizes individual success across robots, and CSR captures the joint outcome of each pair.

\textbf{Shared information and qualitative examples.}
In Table~\ref{tab:share_what}, SEER achieves higher ISR and CSR than the evaluated history-, action-, observation-, and spatial-description-sharing alternatives. Together with the inverse dynamics results in Table~\ref{tab:inverse_dynamics_analysis}, these comparisons connect action-relevant latent structure with downstream navigation performance. The following visualizations show paired observations, language instructions, and executed trajectories across the evaluated spatial configurations.

\begin{figure}[p]
    \centering
    \includegraphics[
        width=\textwidth,
        height=0.86\textheight,
        keepaspectratio,
    ]{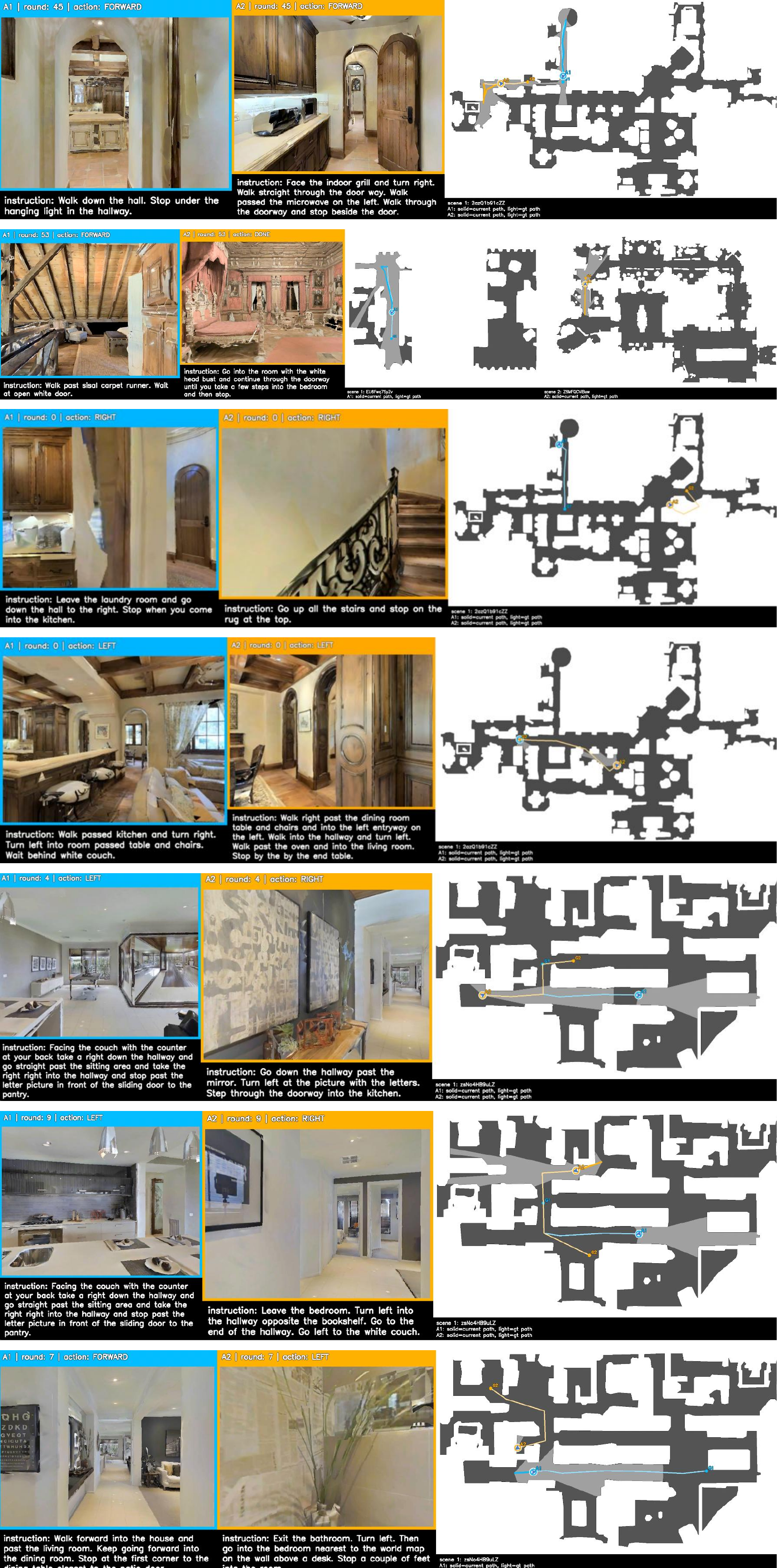}
\end{figure}

\begin{figure}[p]
    \centering
    \includegraphics[
        width=\textwidth,
        height=0.86\textheight,
        keepaspectratio,
    ]{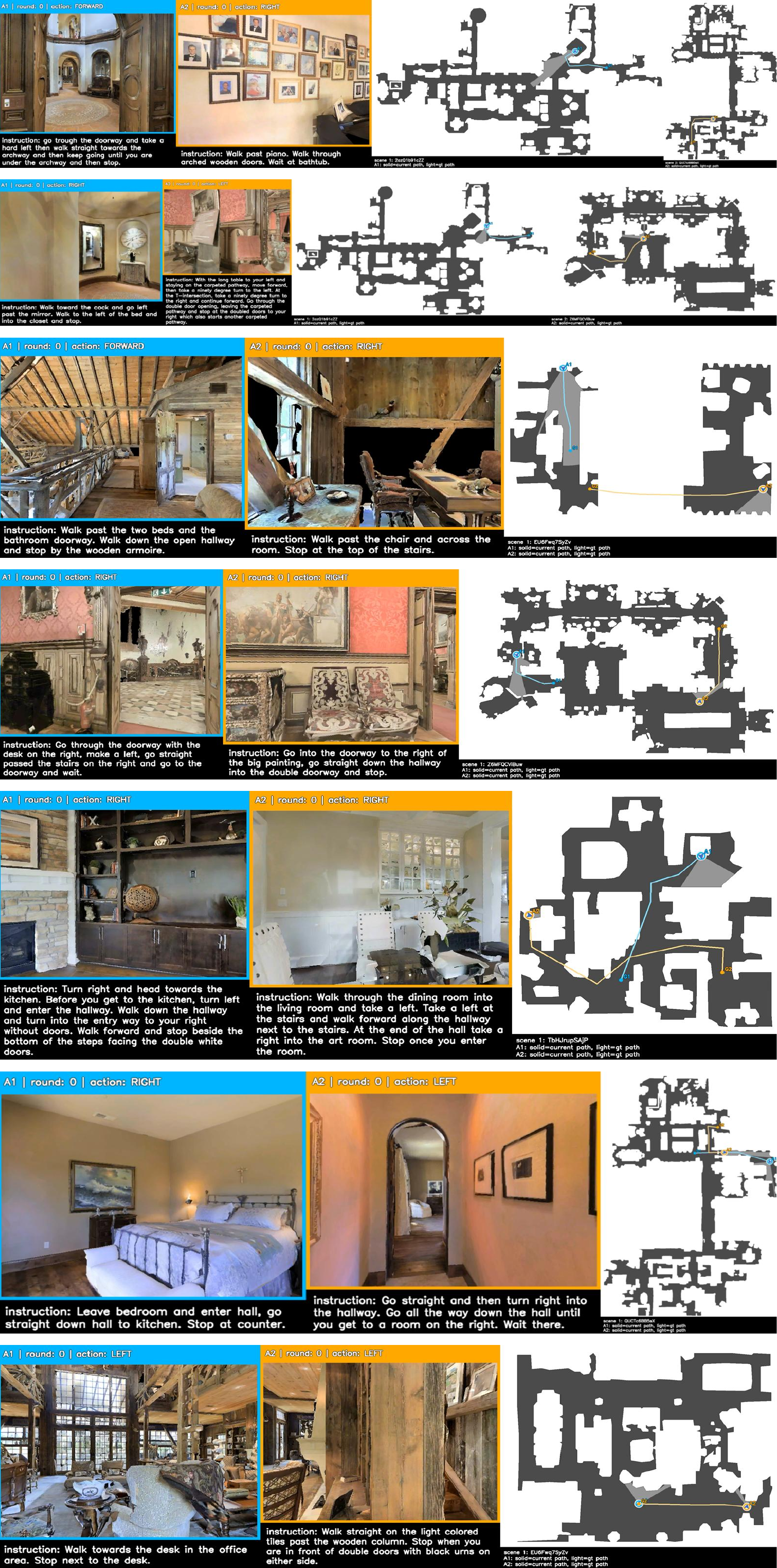}
    \caption{More Qualitative Analysis of our SEER.}
    \label{fig:appendix_test}
\end{figure}